\pdfoutput=1

\documentclass[11pt]{article}

\usepackage[final]{acl}

\usepackage{times}
\usepackage{latexsym}
\usepackage{amsmath}
\usepackage[T1]{fontenc}

\usepackage[utf8]{inputenc}

\usepackage{microtype}

\usepackage{inconsolata}

\usepackage{bm}
\usepackage{amssymb}
\usepackage{xcolor}
\usepackage{float}
\usepackage{booktabs}
\usepackage{multirow}
\usepackage{makecell}

\usepackage{colortbl}
\usepackage{pifont}
\usepackage{graphicx}
\usepackage{subcaption}
\usepackage{algorithm}
\usepackage{algorithmicx}
\usepackage{algpseudocode}
\usepackage{fdsymbol}
\usepackage{pdfpages}

\newcommand{\myalgname}{DTGs\ }
\newcommand{\myalgnamesingle}{DTG\ }
\newcommand{\myalgnamesuffix}{DTG}

\newcommand{\algname}{\textit{arc-swift}}
\newcommand{\arceager}{\textit{arc-eager}}
\newcommand{\arcstandard}{\textit{arc-standard}}
\newcommand{\archybrid}{\textit{arc-hybrid}}

\newcommand{\Arcstandard}{\textit{Arc-standard}}

\title{Dependency Transformer Grammars: Integrating Dependency Structures into Transformer Language Models}

\author{Yida Zhao, Chao Lou, Kewei Tu\thanks{\; Corresponding Author}\\
  School of Information Science and Technology, ShanghaiTech University \\
  Shanghai Engineering Research Center of Intelligent Vision and Imaging\\ 
    {\tt \{zhaoyd2023,louchao,tukw\}@shanghaitech.edu.cn}\\
 }

\begin{document}
\maketitle
\begin{abstract}
Syntactic Transformer language models aim to achieve better generalization through simultaneously modeling syntax trees and sentences. While prior work has been focusing on adding constituency-based structures to Transformers, we introduce Dependency Transformer Grammars (\myalgnamesuffix s), a new class of Transformer language model with explicit dependency-based inductive bias. \myalgname simulate dependency transition systems with constrained attention patterns by modifying attention masks, incorporate the stack information through relative positional encoding, and augment dependency arc representation with a combination of token embeddings and operation embeddings. When trained on a dataset of sentences annotated with dependency trees, \myalgname achieve better generalization while maintaining comparable perplexity with Transformer language model baselines. \myalgname also outperform recent constituency-based models, showing that dependency can better guide Transformer language models. Our code is released at {\url{https://github.com/zhaoyd1/Dep_Transformer_Grammars}}.
\end{abstract}

\section{Introduction}

Transformer language models have shown strong performance on language modeling tasks and a broad spectrum of downstream tasks~\cite{radford2019language, devlin-etal-2019-bert, NEURIPS2020_1457c0d6}.
Despite the great power of the Transformer architecture~\cite{vaswani2017attention}, it lacks the inductive biases of syntactic structures, which has been hypothesized to improve generalization~\cite{everaert2015structures}. 
A straightforward way to incorporate such biases into Transformers is explicit modeling of syntactic structures.

Inspired by earlier work of generative parsing as language modeling that integrates syntactic structures into RNNs~\cite{dyer-etal-2016-recurrent, choe-charniak-2016-parsing}, recent studies have focused on adapting this method to Transformer architectures~\cite{qian-etal-2021-structural, yoshida-oseki-2022-composition, sartran-etal-2022-transformer, murty-etal-2023-pushdown}. The models proposed by these studies are categorized as syntactic language models because they jointly model the distribution of surface strings and their corresponding syntactic trees. Experiments show that these models achieve competitive perplexity in language modeling and gain better syntactic generalization,
supporting the above hypothesis on the benefits of introducing inductive bias of syntactic structures.
However, the structural supervision that has been used in all these models is based on constituency trees and it is unclear of the performance of dependency-based Transformer syntactic language models. Different from constituency structures, which model recursive syntactic compositions, dependency structures focus more on the relationship between tokens, which is similar to the self-attention mechanism in Transformer, hinting at potential synergy between the two. 

In this paper, we propose Dependency Transformer Grammars (\myalgnamesuffix s), dependency-based syntactic language models that learn joint distributions of sentences and dependency trees. \myalgname introduce an inductive bias of dependency structures to Transformers by (i) modeling transition sequences of transition-based dependency parsers instead of sentences, (ii) simulating the stack operations in transition-based dependency parsers through modification of attention masks,
(iii) incorporating the stack information of transition-based systems through relative positional encoding of stack depth, and (iv) representing head-dependent relations through a combination of head token embeddings and transition operation embeddings. Following a line of previous work in generative dependency parsing~\cite{titov-henderson-2007-latent,cohen-etal-2011-exact, buys-blunsom-2015-generative}, the generative formulation of our model is based on the \textit{arc-standard} system~\cite{nivre-2004-incrementality}, which builds a dependency tree in a bottom-up manner. We also explore models using other dependency transition systems for comparison. 


Our experiments show that \myalgname achieve comparable perplexity in language modeling and improved syntactic generalization on both the BLiMP benchmark~\cite{warstadt-etal-2020-blimp-benchmark} and the SG test suites~\cite{hu-etal-2020-systematic} over Transformer language model baselines. Furthermore, \myalgname outperform constituency-based syntactic language models in both language modeling and syntactic generalization. 

In summary, our contributions are as follows.
\begin{itemize}
    \item We propose dependency-based syntactic language models, \myalgnamesuffix s, to incorporate dependency inductive bias into Transformers. 
    \item We primarily build \myalgname using the \textit{arc-standard} transition system, while we also study the usage of other dependency transition systems.
    \item Experimental results on two syntactic generalization benchmarks show the benefits of introducing inductive bias of dependency structures.
\end{itemize}


\begin{table*}[tb]
	\centering
 \resizebox{0.85\textwidth}{!}{%
	\begin{tabular}{@{}ll@{}}
		\toprule
		\arcstandard & \archybrid \\
		$\begin{array}{@{}ll@{}}
		\textbf{Shift} & (\sigma, i|\beta, A) \Rightarrow (\sigma|i, \beta, A) \\
		\textbf{LArc} & (\sigma|i|j, \beta, A) \Rightarrow (\sigma|j, \beta, A \cup \{(j\to i)\}) \\
		\textbf{RArc} & (\sigma|i|j, \beta, A) \Rightarrow (\sigma|i, \beta, A\cup\{(i\to j)\})
		\end{array}$ & 
		$\begin{array}{@{}ll@{}}
		\textbf{Shift} & (\sigma, i|\beta, A)  \Rightarrow  (\sigma|i, \beta, A) \\
		\textbf{LArc} & (\sigma|i, j|\beta, A)  \Rightarrow  (\sigma, j|\beta, A \cup \{(j\to i)\}) \\
		\textbf{RArc} & (\sigma|i|j, \beta, A)  \Rightarrow  (\sigma|i, \beta, A\cup\{(i\to j)\})
		\end{array}
		$
		\\
		\midrule
		\arceager & \algname \\
		$\begin{array}{@{}ll@{}}
		\textbf{Shift} & (\sigma, i|\beta, A)  \Rightarrow  (\sigma|i, \beta, A) \\
		\textbf{LArc} & (\sigma|i, j|\beta, A)  \Rightarrow (\sigma, j|\beta, A \cup \{(j\to i)\}) \\
		\textbf{RArc} & (\sigma|i, j|\beta, A)  \Rightarrow (\sigma|i|j, \beta, A\cup\{(i\to j)\}) \\
		\textbf{Pop} & (\sigma|i, \beta, A)  \Rightarrow  (\sigma, \beta, A)
		\end{array}$ & $\begin{array}{@{}ll@{}}
		\textbf{Shift} & (\sigma, i|\beta, A)  \Rightarrow  (\sigma|i, \beta, A) \\
		\textbf{LArc}[k] & (\sigma|i_k|\ldots|i_1, j|\beta, A) \\
		&\qquad \Rightarrow  (\sigma, j|\beta, A \cup \{(j\to i_k)\}) \\
		\textbf{RArc}[k] & (\sigma|i_k|\ldots|i_1, j|\beta, A) \\ 
		& \qquad \Rightarrow (\sigma|i_k|j, \beta, A\cup\{(i_k\to j)\})
		\end{array}
		$\\
		\bottomrule
	\end{tabular}%
 }
	\caption{Transitions defined by different transition systems (adapted from \citet{qi-manning-2017-arc})} \label{tab:transsys}
\end{table*}

\section{Preliminaries: Transition-based Dependency Parsing}

Given a sentence, transition-based dependency parsing predicts a sequence of predefined transitions between states that incrementally build a dependency parse tree. A state contains a \textit{stack} $\sigma$ with token $i$ on the top, a \textit{buffer} $\beta$ with $j$ at its leftmost side, and a set $A$ of dependency arcs, denoted as $(\sigma|i, j|\beta, A)$. 

In this work, we focus on unlabeled projective dependency parsing for the simplicity of its transition systems.
There are several different transition systems for projective dependency parsing, as shown in Table~\ref{tab:transsys}. \textit{Arc-standard}~\cite{nivre-2004-incrementality} is a widely used transition system that defines three transitions: \texttt{SHIFT}, \texttt{LEFTARC} and \texttt{RIGHTARC}. \Arcstandard\ builds dependency trees in a bottom-up manner, that is, every token is not connected to its head token until it gathers all of its dependents. \textit{Arc-eager}~\cite{nivre-2003-efficient} is another transition system that adds one more transition: \texttt{POP}. The main difference between \textit{arc-standard} and \textit{arc-eager} lies in the scope of arcs. \textit{Arc-standard} only allows inducing arcs in the stack while \textit{arc-eager} eases the restriction by defining arc transitions between the stack and the buffer. As a result, dependency trees are no longer built from bottom to up in \textit{arc-eager}. A later system \textit{arc-hybrid}~\cite{kuhlmann-etal-2011-dynamic} combines \texttt{LEFTARC} in \textit{arc-eager} and \texttt{RIGHTARC} in \textit{arc-standard}. Another more recent system \textit{arc-swift}~\cite{qi-manning-2017-arc} extends arc-inducing to non-local cases: transition \texttt{LEFTARC/RIGHTARC}[$k$] in \textit{arc-swift} can be seen as $k-1$ \texttt{POP} operations followed by one arc-inducing in \textit{arc-eager}. 

The above dependency parsing transition systems can be changed into a generative form, such that they generate sentences along with their associated dependency trees. The main change to the transition systems is that tokens need to be generated instead of being shifted from the buffer. Specifically, in \arcstandard\ we substitute \texttt{SHIFT} with a token generation transition \texttt{GEN}, while retaining the other transitions~\cite{titov-henderson-2007-latent,cohen-etal-2011-exact, buys-blunsom-2015-generative}. Other systems require additional efforts to obtain a generative form because they contain the usage of the buffer head in \texttt{LEFTARC} and/or \texttt{RIGHTARC} before shifting it to the stack. Simply replacing \texttt{SHIFT} with \texttt{GEN} cannot ensure the existence of the two tokens involved in a newly generated arc. Therefore, we need to insert a \texttt{GEN'} transition,\footnote{To simplify, we will refer to \texttt{GEN'} as \texttt{GEN}, which can be distinguished according to transition systems.} which generates a new token but puts it in the buffer, before any \texttt{LEFTARC/RIGHTARC} transition that involves an ungenerated token. The \texttt{SHIFT} transitions are omitted because any generated token will be shifted to the stack once a new token is generated.

We can use an oracle to extract a transition sequence from a dependency parse tree: An arc-inducing transition is generated whenever possible, and a \texttt{POP} transition (in \textit{arc-eager}) is generated when it is impossible to generate other transitions, i.e., the transition preference order is \texttt{LEFTARC/RIGHTARC} > \texttt{GEN} > \texttt{POP}.

\section{Model}

\begin{figure}[tb]
    \centering
    \includegraphics[width=\columnwidth]{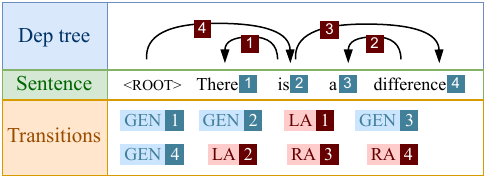}
    \caption{An example sentence with its dependency tree and transition sequence. Numbers in blue and red are indices of tokens and arcs respectively.}
    \label{fig:seq}
\end{figure}

\begin{figure*}[tb]
\centering
\begin{subfigure}[b]{0.485\textwidth}
\centering
\resizebox{\textwidth}{!}{%
\begin{tabular}{c|l|l|l}\toprule
    $i$ & \thead{Input} & \thead{Attn. Mask} & \thead{Prediction} \\\midrule
    0 & <ROOT> & STACK & \texttt{GEN}(There) \\
    1 & There & STACK & \texttt{GEN}(is) \\
    2 & is & STACK & \texttt{LEFTARC} \\
    3 & \texttt{LEFTARC} + is & COMPOSE & - \\
    4 & \texttt{LEFTARC2} + is & STACK & \texttt{GEN}(a) \\
    5 & a & STACK & \texttt{GEN}(difference) \\
    6 & difference  & STACK & \texttt{LEFTARC}\\
    7 & \texttt{LEFTARC} + difference & COMPOSE & - \\
    8 & \texttt{LEFTARC2} + difference & STACK & \texttt{RIGHTARC} \\
    9 & \texttt{RIGHTARC} + is &COMPOSE & - \\
    10 & \texttt{RIGHTARC2} + is & STACK & \texttt{RIGHTARC} \\
    11 & \texttt{RIGHTARC} + <ROOT> & COMPOSE & - \\
    12 & \texttt{RIGHTARC2} + <ROOT> & STACK & <END> \\\bottomrule
\end{tabular}%
}
\bigskip
\caption{Transition sequence after duplicating \texttt{LEFTARC/RIGHTARC} transitions. We do not have to make predictions for positions 3, 7, 9, 11.}
\label{fig:arc_standard_seq_and_mask_subfig_seq}
\end{subfigure}
\hfill
\begin{subfigure}[b]{0.485\textwidth}
    \centering
    \includegraphics[width=\textwidth,clip, trim=2.4cm 0cm 0cm 0cm]{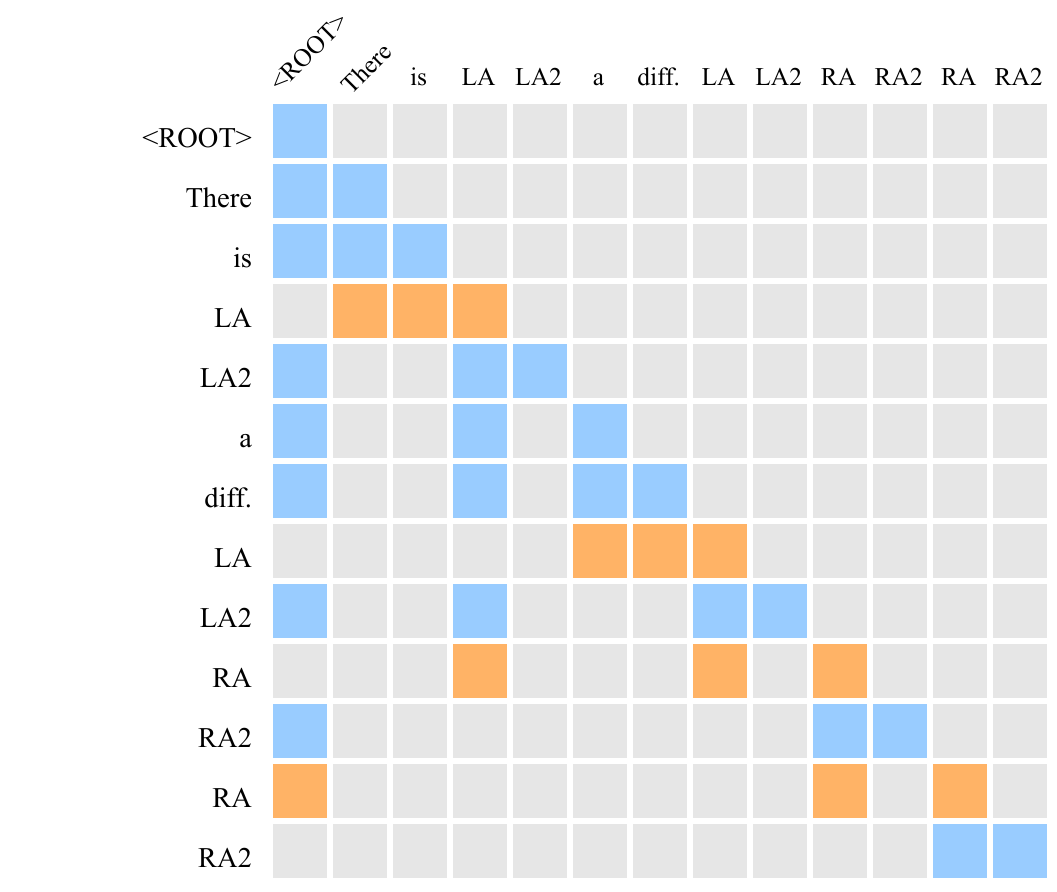}
    \caption{Attention mask. Tokens are simplified for a tight view. We use orange to represent \textbf{COMPOSE} and blue to represent \textbf{STACK}.}
    \label{fig:arc_standard_seq_and_mask_subfig_mask}
\end{subfigure}
\caption{Transition sequence and attention masks of an example sentence}
\label{fig:arc_standard_seq_and_mask}
\end{figure*}

\myalgnamesingle follows the generative form of the \textit{arc-standard} dependency transition system and generates a sequence of transitions that construct a sentence $\mathbf{x}$ and its dependency tree $\mathbf{y}$ incrementally. The sequence consists of three types of transitions: 
\begin{itemize}
    \item \textbf{\texttt{GEN}$\bm{(}x\bm{)}$}: generating a token $x$, which corresponds to the \texttt{GEN} operation in generative \textit{arc-standard} and is exactly what a standard Transformer decoder does at each step;
    \item \textbf{\texttt{LEFTARC}} or \textbf{\texttt{LA}}: inducing an arc from the most recent unconnected token (i.e., a token that has not been connected to its head) to the second most recent unconnected token, which corresponds to the \texttt{LEFTARC} operation in \textit{arc-standard};
    \item \textbf{\texttt{RIGHTARC}} or \textbf{\texttt{RA}}: inducing an arc from the second most recent unconnected token to the most recent unconnected token, which corresponds to the \texttt{RIGHTARC} operation in \textit{arc-standard}. 
\end{itemize}
An example is shown in Figure \ref{fig:seq}.

We write $\boldsymbol{\alpha}(\mathbf{x}, \mathbf{y}) = (\alpha_0, \alpha_1, ..., \alpha_{T-1})$ as the transition sequence of length $T$ of sentence $\mathbf{x}$ and parse tree $\mathbf{y}$, where each $\alpha_t$ belongs to one of the three types mentioned above. \myalgnamesingle is a Transformer decoder that models the distribution of $\boldsymbol{\alpha}(\mathbf{x}, \mathbf{y})$ in the manner of causal language modeling,
that is, $p(\boldsymbol{\alpha}(\mathbf{x}, \mathbf{y})) = \prod\limits_i p(\alpha_i | \boldsymbol{\alpha}_{<i})$. It differs from a standard Transformer in several aspects in order to incorporate the dependency inductive bias, including attention masks, positional encoding, augmented representation of arcs, and constrained generation, which we discuss in the following subsections.

\subsection{Arc-Standard via Attention Mask}
\label{sec:arc-standard-via-attention}



DTGs generate the transition sequence autoregressively. A standard Transformer language model makes predictions based on the complete generation history. In contrast, to incorporate the dependency inductive bias into DTGs, we generate transitions based on the stack in \textit{arc-standard}. The stack is encoded into the model with different attention forms and is updated by input transitions.

When a \texttt{GEN} transition comes, the transition system pushes a new token onto the stack and then gathers the stack information to generate the next transition, which we realize by the first attention form, \textbf{STACK} attention.
When a transition changing the dependency structure comes, i.e., a \texttt{LEFTARC/RIGHTARC} transition, the stack is updated in two steps: (i) pop two tokens from the stack and designate one as the head of the other and (ii) push the head token back onto the stack. The two steps are realized by the second form of attention, \textbf{COMPOSE} attention, which updates the representation of the head by
consuming its dependent but ignoring everything
else to reflect the newly induced dependency arc. Then all the stack information is gathered for generating the next transition, which is again realized by \textbf{STACK} attention. 
Therefore, two forms of attention are required for one transition. 
As each transition can only use one form of attention in Transformer,
we duplicate the arc transitions, namely \texttt{LEFTARC/RIGHTARC} and \texttt{LEFTARC2/RIGHTARC2}. The former encodes dependency information with \textbf{COMPOSE} attention and makes no generation, while the latter triggers the generation of the next transition with \textbf{STACK} attention. After the duplication, the sequence length increases from $T$ to $T'$. We denote the new sequence as $\boldsymbol{\alpha}'$, which is the exact input sequence of our model. Note that this does not change the distribution of $\boldsymbol{\alpha}(\mathbf{x}, \mathbf{y})$, as the generation sequence remains unchanged.
An example of the expanded transition sequence and the corresponding attention forms is shown in Figure~\ref{fig:arc_standard_seq_and_mask_subfig_seq}. 

\begin{algorithm}[tb]
\begin{small}
\caption{\textsc{compose}/\textsc{stack} attention}
\label{stack/compose attention}
\begin{algorithmic}[1]

\Require $\boldsymbol{\alpha}'$ sequence of transitions
\Ensure $\mathbf{A} \in \mathbb{R}^{T' \times T'}$ attention mask
\State $S \gets []$\Comment{Empty stack}
\State $\mathbf{A} \gets 0$
\For{$i \gets 0$ to $T'$}
    \If{$\textrm{type}(\boldsymbol{a}'[i]) = \texttt{LEFTARC}$ or
        \State  $\textrm{type}(\boldsymbol{a}'[i]) = \texttt{RIGHTARC}$}
        \Comment{\textsc{compose}}
        \State $A_{ii} \gets 1$
        \State l$ \gets S.pop()$ 
        \State r$ \gets S.pop()$
        \State $A_{il} \gets 1$
        \State $A_{ir} \gets 1$
        \State $S.push(i)$ \Comment{View transition $i$ as the head token}
    \Else \Comment{\textsc{stack}}
        \If{$\textrm{type}(\boldsymbol{a}'[i]) \neq \texttt{LEFTARC2}$ and
        \State  $\textrm{type}(\boldsymbol{a}'[i]) \neq \texttt{RIGHTARC2}$}
        \State $S.push(i)$ 
        \EndIf
        \For{$j \in S$}
            \State $A_{ij} \gets 1$
        \EndFor
    \EndIf
\EndFor
\State \textbf{return} $\mathbf{A}$\Comment{Attention mask}
\end{algorithmic}\end{small}
\end{algorithm}

The two forms of attention can be realized by leveraging different attention masks. We represent the attention masks as $\mathbf{A} \in \mathbb{R}^{T' \times T'}$, where $A_{ij} = 1$ means position $j$ can be attended from $i$ and $A_{ij} = 0$ means position $j$ is masked from $i$. Our models generate transitions in an autoregressive manner, so the attention mask is causal, i.e., $A_{ij} = 0 \ \text{for} \ j > i$. 

\textbf{STACK} attention is performed at each position $i$ which needs to predict a new transition, i.e., $\alpha'_i \in \{\texttt{GEN}(x), \texttt{LEFTARC2}, \texttt{RIGHTARC2}\}$. From position $i$, we attend to all the unmasked positions before $i$ (including $i$) to collect all the information on the stack for generation.

\textbf{COMPOSE} attention is performed at each position $i$ where $\alpha'_i \in \{\texttt{LEFTARC}, \texttt{RIGHTARC}\}$. From position $i$, we attend to the positions of the most recent two unmasked tokens, i.e., the top two tokens on the stack in \textit{arc-standard}, which forms a head-dependent pair. Then we mask the two attended positions from subsequent positions, effectively popping the two tokens from the stack. The newly computed representation serves as a substitute for the head token that has absorbed the information of its dependent and is pushed back onto the stack. 

Algorithm~\ref{stack/compose attention} shows how to compute attention masks for a transition sequence as described above.
We also show attention masks of an example transition sequence in Figure~\ref{fig:arc_standard_seq_and_mask_subfig_mask}.

\subsection{Relative Positional Encoding}

We design the positional encoding for \myalgname based on the relative positional encoding in Transformer-XL~\cite{dai-etal-2019-transformer}. In Transformer-XL, the positional encoding is based on the distance between the attending position $i$ and the attended position $j$, i.e., $\mathbf{R}_{ij} = i-j$. In \myalgnamesuffix s, we modify the formulation to reflect the stack information. Crucially, $\mathbf{R}_{ij}$ is only computed when $A_{ij} = 1$. For \textbf{STACK} attention, we define $d(i)$ as the depth in the stack, which increases from the top to the bottom. We then define $\mathbf{R}_{ij} = d(i) - d(j)$. For \textbf{COMPOSE} attention, we define two positions, $0$ and $-1$, to distinguish between the head and the dependent to be composed, i.e., $\mathbf{R}_{ij} = 0$ if token $j$ is the head token and $\mathbf{R}_{ij} = -1$ if token $j$ is the dependent token. The new representation computed with \textbf{COMPOSE} inherits the depth of the head token, i.e., $d(i) = d(j)$ if token $j$ is the composed head token.

\subsection{Arc Representation}

In standard language models, generated tokens are fed back into models as history. For arc-inducing transitions in \myalgnamesuffix s, the generated transitions have surface forms of \texttt{LEFTARC} or \texttt{RIGHTARC} while the tokens ought to be pushed back are the head tokens. 
We propose to feed a combination of \texttt{LEFTARC/RIGHTARC} and the head token via summing the embedding of these two parts.
This formulation stems from the following two considerations: (i) the attention in \myalgname cannot distinguish between \texttt{LEFTARC} and \texttt{RIGHTARC}, so the embedding of \texttt{LEFTARC/RIGHTARC} acts as an indicator of the arc direction; (ii) the representation computed with \textbf{COMPOSE} is viewed as a substitute of the composed head token by subsequent positions, so we add the embedding of the head token to bias the representation.

\subsection{Other Transition Systems via Attention Mask}

We also design the attention mechanism for generative \textit{arc-eager} and \textit{arc-swift} and name the resulting models \myalgnamesuffix-eager and \myalgnamesuffix-swift. We do not work on generative \textit{arc-hybrid} because its transition sequences are exactly the same as that of generative \textit{arc-standard}.

For \myalgnamesuffix-eager, we make two modifications based on \myalgnamesuffix: 
    (i) Change the \textbf{COMPOSE} attention of \texttt{RIGHTARC} by not masking the position of the dependent token because in \textit{arc-eager}, the dependent token can still induce arcs to subsequent tokens.
    (ii) For transition \texttt{POP}, we define \textbf{POPSTACK} attention, which pops the stack top. The stack top is the second most recent unmasked token in most cases, and the most recent one is the head of the buffer. However, if all tokens have been generated and thus the buffer is empty, the stack top is the most recent unmasked token.

For \myalgnamesuffix-swift, \texttt{LEFTARC} and \texttt{RIGHTARC} are decorated by an additional positive number $k$. This affects ranges of attending and masking in \textbf{COMPOSE} attention. That is, we attend to not only the head-dependent pair but also the $k-1$ tokens between them, and we mask all these $k+1$ tokens for subsequent positions.

More details and examples of these two models are provided in Appendix \ref{sec:appendix}.

\subsection{Constraints on Inference}

We define several constraints on transition generation during \myalgname inference to make it consistent with the corresponding transition-based dependency parsing systems:
\begin{itemize}
    \item For all the systems, the \texttt{LEFTARC} and \texttt{RIGHTARC} transition can only be generated if at least two tokens exist in the stack.
    \item For \textit{arc-eager}, \texttt{POP} can only be generated if the top of the stack has been recognized as a right dependent of some head token. 
    \item For \textit{arc-swift}, the value of $k$ in \texttt{LEFTARC/RIGHTARC}[$k$] must not exceed the size of the stack.   
\end{itemize}

\section{Experiments} \label{Experiments}

We compare \myalgname with \myalgnamesuffix-eager, \myalgnamesuffix-swift, two Transformer-XL baselines, and constituency-based syntactic Transformer language models. The two Transformer-XL baselines follow those of \citet{sartran-etal-2022-transformer}: (i) \textbf{TXL (tokens)} is a standard Transformer-XL that generates sentences only, and (ii) \textbf{TXL (trans)} is Transformer-XL that generates transition sequences just like DTG, but uses standard attention masks and positional encoding. Constituency-based syntactic Transformer language models include: (i) the ``generative
parsing as language modeling'' of \citet{qian-etal-2021-structural} (\textbf{PLM}), (ii) Transformer Grammars of \citet{sartran-etal-2022-transformer} (\textbf{TG}) and (iii) Pushdown Layers of \citet{murty-etal-2023-pushdown} (\textbf{Pushdown}). 

\paragraph{Dataset and Preprocessing}

All the models are trained on the BLLIP{\small-LG} dataset of \citet{BLLIP-2000}, with training splits from \citet{hu-etal-2020-systematic}. For our models, we obtain unlabeled projective dependency trees by parsing the dataset with a Biaffine{\small-roberta} parser~\cite{dozat2017deep} implemented in \textit{Supar}\footnote{\url{https://github.com/yzhangcs/parser}}. Tokenization is performed with the same scheme as in \citet{sartran-etal-2022-transformer} with SentencePiece~\cite{kudo-richardson-2018-sentencepiece}. Note that we model each sentence independently in all the experiments.

\paragraph{Training Details} 

We use the same hyperparameters as in \citet{sartran-etal-2022-transformer} for training our models, using 16-layer models with 252M parameters. To accelerate the training of token embeddings, we add a multiplier of 2.0 to the learning rate of embedding weights. More details can be found in Appendix~\ref{sec:other_exp_details}.

\begin{table}[tb]
    \centering
    \resizebox{\columnwidth}{!}{%
    \begin{tabular}{l|l|ccc} \toprule
        \multicolumn{2}{c}{\thead{Model}} & \thead{PPL ($\downarrow$)} & \thead{BLiMP ($\uparrow$)} & \thead{SG ($\uparrow$)} \\\midrule
        \multicolumn{5}{c}{\cellcolor{gray!20}\textit{Models without syntactic inductive bias}} \\
        \multicolumn{2}{l|}{TXL (tokens)} & $14.8$ & $75.3$ & $76.6$\\
        \multicolumn{5}{c}{\cellcolor{gray!20}\textit{Constituency-based models}} \\
         \multicolumn{2}{l|}{PLM} & $29.8$\rlap{$^\diamondsuit$} & $75.1$ & $80.2$\\
         \multicolumn{2}{l|}{TG}  & $18.4$\rlap{$^\clubsuit$} & $73.5$\rlap{$^\clubsuit$} & $82.5$ \\
         \multicolumn{2}{l|}{Pushdown} & $19.9$\rlap{$^\diamondsuit$} & $75.6$ &  $82.3$\\
         \multicolumn{5}{c}{\cellcolor{gray!20}\textit{Dependency-based models}} \\
         \multicolumn{2}{l|}{TXL (trans)} & $\mathbf{14.4}$ & $\mathbf{77.3}$ & $81.1$\\
         \multirow{3}{*}{\rotatebox[origin=c]{90}{Ours}}& \myalgnamesuffix-eager & $15.5$ & $75.2$ & - \\
         & \myalgnamesuffix-swift & $15.0$ & $76.2$ & - \\
         & \myalgnamesuffix & $14.9$ & $76.1$ & $\mathbf{83.9}$ \\ \bottomrule
    \end{tabular}%
    }
    \caption{Results of our models and baselines. $\diamondsuit$: Results are taken from prior work and are only for reference due to differences in tokenization. $\clubsuit$: We rerun the code from the original work~\cite{sartran-etal-2022-transformer} and obtain better perplexity than the reported result in it. All results for PLM and Pushdown are taken from \citet{murty-etal-2023-pushdown}. The SG result for TG is taken from \citet{sartran-etal-2022-transformer}.}
    \label{tab:sg_resuls}
\end{table}

\subsection{Sentence-Level Language Modeling}

We evaluate the perplexity of the models on the BLLIP{\small-LG} dataset of \citet{BLLIP-2000}, with test splits from \citet{hu-etal-2020-systematic}.
\paragraph{Setup}\label{4.1}

For syntactic language models that jointly model the distributions of sentences and syntactic trees, i.e., $p(\mathbf{x},\mathbf{y})$, we compute the string probability $p(\mathbf{x}) = \sum_{\mathbf{y}}p(\mathbf{x},\mathbf{y})$. It is impossible to compute $p(\mathbf{x})$ precisely due to the large space of all possible trees, so we follow \citet{sartran-etal-2022-transformer} to approximate it using a relatively small set of trees sampled from a proposal model $q(\mathbf{y} | \mathbf{x})$. For our depdendency-based models, we use the Biaffine{\small-roberta}~\cite{dozat2017deep} parser as the proposal model to sample 300 unlabeled projective dependency trees without replacement as a proposal tree set $\mathbf{Y'}$. $p(\mathbf{x})$ is then approximated by $\sum_{\mathbf{y} \in \mathbf{Y'}}p(\mathbf{x},\mathbf{y})$, which is an exact lower bound of the true value of $p(\mathbf{x})$ (hence leading to an upper bound of perplexity). We evaluate the models by sentence-level perplexity.

\paragraph{Results}

We report the perplexity of all the models in Table \ref{tab:sg_resuls}. \myalgnamesingle achieves comparable perplexity with TXL (tokens) and \myalgnamesuffix-swift, outperforming \myalgnamesuffix-eager. 
TXL (trans) achieves lower perplexity than TXL (tokens) even though the reported result of TXL (trans) is an upper bound of its true perplexity. It shows that jointly modeling dependency trees and sentences is helpful for sentence-level language modeling.

The perplexity upper bound of \myalgnamesingle can be seen to be lower than that of TG. There are two possible interpretations of this result: (i) Dependency trees give better guidance than constituency trees in syntactic language modeling. (ii) 300 trees may be too few to get an accurate approximation of perplexity when sampling from a large set of possible trees. Evaluating \myalgnamesingle and TG requires samples of unlabeled projective dependency trees and labeled constituency trees, respectively. The number of the former is much smaller than the number of the latter. Therefore, sampling 300 trees may give a much tighter perplexity upper bound for \myalgnamesingle than for TG, resulting in a gap in the reported results.
Unfortunately, it requires nontrivial work to distinguish between the two possibilities and we leave it for future work.

\subsection{Syntactic Generalization}

\begin{figure}
    \centering
    \includegraphics[width=\columnwidth,clip,trim=0.6cm 0.6cm 0cm 0cm]{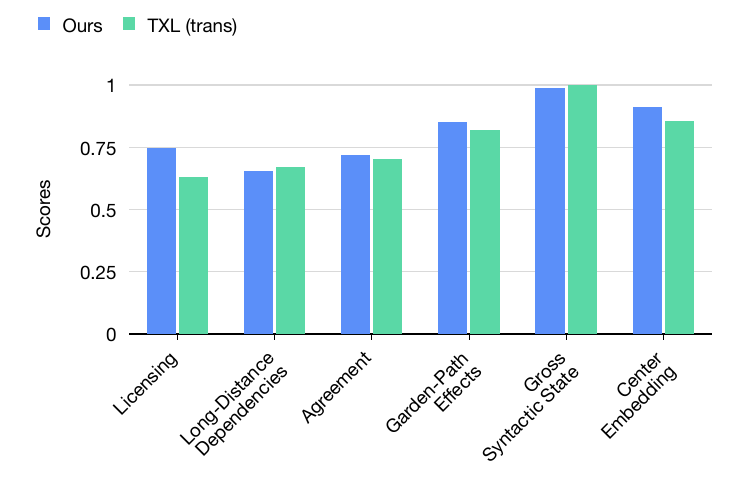}
    \caption{Scores on the six circuits of the SG test suites.}
    \label{fig:circuit}
\end{figure}

To measure the syntactic generalization, we evaluate our models on BLiMP~\cite{warstadt-etal-2020-blimp-benchmark} and SG test suites~\cite{hu-etal-2020-systematic}.

\paragraph{Setup on BLiMP}

BLiMP contains 67 generalization tests, each with 1000 sentence pairs. Each sentence pair consists of a grammatical sentence and an ungrammatical sentence. Models are evaluated by whether they assign a higher probability to the grammatical one.  We use the same setup as in Section \ref{4.1}, sampling 300 trees for each sentence and calculating a lower bound of marginal probability $p(\mathbf{x})$ for comparison.

\paragraph{Setup on SG}

SG consists of test suites for six fine-grained syntactic phenomena. Each test suite has a specific inequality formula for evaluation. These inequalities are based on incremental natural processing, requiring computing the surprisal values, i.e., $-\log p(x_t | x_{<t})$. We implement the word-synchronous beam search~\cite{stern-etal-2017-effective,hale-etal-2018-finding} to get the marginal probability at each token $t$ and calculate the surprisal value. We fix the beam size at 300. 

\paragraph{Results}

The results are reported in Table \ref{tab:sg_resuls}. For BLiMP, we found that most of the constituency-based syntactic language models perform comparably with our baseline TXL (tokens), while \myalgnamesuffix, \myalgnamesuffix-swift, and TXL (trans) outperform them. For SG, all syntactic language models perform better than TXL (tokens), and \myalgnamesingle achieves the highest score. 
These results show that explicit modeling of syntactic structures is helpful for better generalization in Transformer language models, and dependency relations may lead to greater improvements in generalization than constituency compositions. 

We further compare TXL (trans) with \myalgnamesuffix. The SG scores of 6 circuits are shown in Figure \ref{fig:circuit}. In SG, \myalgnamesingle achieves a much higher average score than TXL (trans) and outperforms TXL (trans) in 4 circuits while maintaining comparable scores in the other 2 circuits. Further discussion of SG scores can be found in Appendix~\ref{sec:discussion_SG}.  
On the other hand, TXL (trans) performs better than \myalgnamesingle on BLiMP. We believe it is because BLiMP evaluates semantic knowledge in addition to syntactic knowledge as detailed in \citet{warstadt-etal-2020-blimp-benchmark}, even though BLiMP is used as a syntactic testset in previous work of syntactic language models \cite{qian-etal-2021-structural,murty-etal-2023-pushdown}. Syntax-motivated attention masking in \myalgnamesuffix, while helpful in syntactic modeling, hinders acquisition of semantic information. Please refer to Appendix~\ref{sec:discussion_blimp} for more discussion. 
It is thus an interesting future direction to integrate syntactic language models with standard language models so as to get the best of both worlds.

\subsection{Parse Reranking}

\paragraph{Setup}

We study to what extent \myalgnamesingle and TXL~(trans) have learned to produce correct dependency structures. We still sample 300 trees with the Biaffine{\small-roberta} parser and rerank them using the two models. We convert human-annotated constituency trees in the Penn Treebank (PTB)~\cite{marcus-etal-1993-building} test split into dependency trees with CoreNLP 3.3.0~\cite{manning-etal-2014-stanford} and then evaluate the UAS of the reranked trees on them.

\begin{table}[tb]
    \centering
    \resizebox{0.5\columnwidth}{!}{%
    \begin{tabular}{l|c} \toprule
        \thead{Model} & \thead{UAS ($\uparrow$)} \\\midrule
         Biaffine{\small-roberta} & 96.9 \\
         TXL (trans) & 97.0 \\
         \myalgnamesuffix & 97.0 \\ \bottomrule    
    \end{tabular}%
    }
    \caption{UAS on the PTB test set.}
    \label{tab:parse}
\end{table}
\paragraph{Result}

We present the results in Table \ref{tab:parse}. TXL (trans) and \myalgnamesingle both achieve a slightly higher score than the proposal model Biaffine{\small-roberta}. Note that both models are trained on the dependency parse trees produced by Biaffine{\small-roberta}. The results show that both models successfully learn about dependency structures from Biaffine{\small-roberta}.

\section{Analysis}

\subsection{Arc Representation}


We compare three different representations of \texttt{LEFTARC/RIGHTARC} in \myalgnamesingle: (i) the default formulation of summing the \texttt{LEFTARC/RIGHTARC} embedding and the embedding of the head token $x$, (denoted as \textit{w + arc}); (ii) the embedding of the \texttt{LEFTARC/RIGHTARC} alone (denoted as \textit{arc}); (iii) the embedding of the head token alone (denoted as \textit{w}). \myalgnamesingle models with these representations are trained and evaluated with the same setting as in Section \ref{Experiments}. 

\begin{table}[tb]
    \centering
    \resizebox{0.64\columnwidth}{!}{%
    \begin{tabular}{l|cc} \toprule
        \thead{Model} & \thead{PPL ($\downarrow$)} & \thead{BLiMP ($\uparrow$)}  \\\midrule
         \textit{w} & $15.1$ & $75.9$\\
         \textit{arc} & $15.2$ & $75.8$\\
         \textit{w+arc} & $\mathbf{14.9}$ & $\mathbf{76.1}$\\ \bottomrule
    \end{tabular}%
    }
    \caption{Results of different arc representations.}
    \label{tab:arc}
\end{table}

The result is reported in Table \ref{tab:arc}. The default formulation outperforms the other two representations, showing that both the head token embedding and the \texttt{LEFTARC/RIGHTARC} embedding play a positive role in arc representation. 

\subsection{Dependency Parses for Training}

We use an external parser to provide dependency trees in the training data and sample 300 trees in sentence probability evaluation. Here, we study how the quality of the external parser affects our model's performance. We compare two parsers, vanilla Biaffine without pre-trained token embeddings and Biaffine{\small -roberta},\footnote{Also from \url{https://github.com/yzhangcs/parser}} as the external parser used in training and evaluation. Note that Biaffine{\small -roberta} is more accurate than vanilla Biaffine.

\begin{table}[tb]
    \centering
    \resizebox{0.8\columnwidth}{!}{%
    \begin{tabular}{l|cc} \toprule
        \thead{Parser} & \thead{PPL ($\downarrow$)} & \thead{BLiMP ($\uparrow$)}  \\\midrule
         Biaffine & $15.1$ & $76.0$\\
         Biaffine{\small -roberta} & $\mathbf{14.9}$ & $\mathbf{76.1}$\\ \bottomrule    
    \end{tabular}%
    }
    \caption{Results of using different external parsers.}
    \label{tab:biaffine}
\end{table}

The result is reported in Table \ref{tab:biaffine}. We see an improvement in both perplexity and generalization when using a better parser. 

\section{Related Work}

Augmenting language models with syntactic bias has been studied for a long time. One line of work adds constituency-based syntactic structures to language models through jointly modeling the distribution of sentences and structures~\cite{chelba-1997-structured, roark-2001-probabilistic, henderson-2004-discriminative,  choe-charniak-2016-parsing, kim-etal-2019-unsupervised}. The RNNG model~\cite{dyer-etal-2016-recurrent} is a representative work of syntactic language models, using recursive networks to build representations of phrases. 
More recent work of syntactic language models is based on Transformers~\cite{qian-etal-2021-structural, yoshida-oseki-2022-composition, sartran-etal-2022-transformer,  murty-etal-2023-pushdown}. \citet{qian-etal-2021-structural} and \citet{sartran-etal-2022-transformer} constrain the attention with syntactic bias, while Pushdown Layers~\cite{murty-etal-2023-pushdown} enforce structural constraints via gradient based learning. The above work is all based on constituency structures, and there has been some work considering dependency trees with simple neural networks~\cite{titov-henderson-2007-latent,cohen-etal-2011-exact, buys-blunsom-2015-generative,mirowski-vlachos-2015-dependency}. Most of them, however, focus more on generative dependency parsing while scratching the surface of a language modeling setting. A more general work is \citet{prange-etal-2022-linguistic}, which both introduces constituency and dependency graphs to augment Transformer language modeling, but it requires given gold trees for generation. Following the work of generative dependency parsing and the constrained attention patterns used in \citet{sartran-etal-2022-transformer} and other work~\cite{strubell-etal-2018-linguistically, peng-etal-2019-palm, zhang2020sg, Nguyen2020Tree-Structured, fernandez-astudillo-etal-2020-transition, lou-tu-2023-amr}, we propose \myalgnamesuffix, a novel class of dependency-based syntactic language models. It is the first syntactic language model that designs a dependency-based constrained attention mechanism for Transformers.

Another line of work augments models by learnable structures. Some studies integrate stack-structured memory into models, where updating patterns are learned from data rather than being dictated by predefined syntactic inductive bias~\cite{joulin2015inferring,yogatama2018memory,dusell2021learning, dusell2023stack}. Besides, some studies propose to learn structural attention patterns~\cite{kim2017structured,wang-etal-2019-tree,shen-etal-2021-structformer,shen-etal-2022-unsupervised}. For example, \citet{kim2017structured} assumes that the attention scores are subject to linear-chain or tree conditional random fields~\cite[CRFs;][]{10.5555/645530.655813}. These kinds of augmentation lead to better generalization but usually cost longer running time than naive counterparts.


Some other studies focus on examining the syntactic knowledge acquired by standard attention after pretraining~\cite{Htut2019DoAH,kovaleva-etal-2019-revealing,Kim2020Are,ravishankar-etal-2021-attention}. These studies have identified that certain attention heads align their attention patterns with syntactic structures, thereby providing substantial evidence for the benefits of introducing syntactic inductive bias. In addition, some work re-invents attention using dependency structures and CRFs~\cite{wu-tu-2023-probabilistic}, motivating more linguistically principled studies.


\section{Conclusion}
We propose \myalgnamesuffix s, a new type of syntactic language models that add explicit dependency bias into Transformers. \myalgname simulate dependency transition systems with constrained attention patterns and incorporate stack information through relative positional encoding. 
Experiments show that \myalgname surpass Transformer language model baselines and other constituency-based syntactic language models on syntactic generalization while maintaining competitive perplexity. This implies that the presence of dependency information does improve the performance of Transformer language models.

\section*{Limitations}

\myalgname rely on dependency trees for training, which are predicted by an external parser in this study. However, for languages lacking accurate dependency parsers, our methods might not offer benefits. Additionally, we restrict trees in our study to be in the Standard Dependency representation~\cite{de-marneffe-manning-2008-stanford} and only consider non-labeled projective dependency trees at the sentence level. The investigation of other dependency representations, such as Universal Dependencies~\cite{Nivre2020UniversalDV}, more complex trees and document-level settings is left for future research. 

For training and inference, \myalgname cannot utilize some recent advancements for Transformers easily, including rotary position embeddings~\cite{Su2021RoFormerET} and Flash attention~\cite{Dao2022FlashAttentionFA}, due to our attention mask patterns and relative position encodings. Moreover, evaluating a sentence's probability with \myalgname requires marginalizing over all possible trees, which is intractable. In this study, we approximate this by sampling 300 trees. However, this is still time-consuming and only provides an upper bound for the perplexity metric.



\bibliography{custom,anthology}

\begin{thebibliography}{58}
\expandafter\ifx\csname natexlab\endcsname\relax\def\natexlab#1{#1}\fi

\bibitem[{Brown et~al.(2020)Brown, Mann, Ryder, Subbiah, Kaplan, Dhariwal, Neelakantan, Shyam, Sastry, Askell, Agarwal, Herbert-Voss, Krueger, Henighan, Child, Ramesh, Ziegler, Wu, Winter, Hesse, Chen, Sigler, Litwin, Gray, Chess, Clark, Berner, McCandlish, Radford, Sutskever, and Amodei}]{NEURIPS2020_1457c0d6}
Tom Brown, Benjamin Mann, Nick Ryder, Melanie Subbiah, Jared~D Kaplan, Prafulla Dhariwal, Arvind Neelakantan, Pranav Shyam, Girish Sastry, Amanda Askell, Sandhini Agarwal, Ariel Herbert-Voss, Gretchen Krueger, Tom Henighan, Rewon Child, Aditya Ramesh, Daniel Ziegler, Jeffrey Wu, Clemens Winter, Chris Hesse, Mark Chen, Eric Sigler, Mateusz Litwin, Scott Gray, Benjamin Chess, Jack Clark, Christopher Berner, Sam McCandlish, Alec Radford, Ilya Sutskever, and Dario Amodei. 2020.
\newblock \href {https://proceedings.neurips.cc/paper_files/paper/2020/file/1457c0d6bfcb4967418bfb8ac142f64a-Paper.pdf} {Language models are few-shot learners}.
\newblock In \emph{Advances in Neural Information Processing Systems}, volume~33, pages 1877--1901. Curran Associates, Inc.

\bibitem[{Buys and Blunsom(2015)}]{buys-blunsom-2015-generative}
Jan Buys and Phil Blunsom. 2015.
\newblock \href {https://doi.org/10.3115/v1/P15-2142} {Generative incremental dependency parsing with neural networks}.
\newblock In \emph{Proceedings of the 53rd Annual Meeting of the Association for Computational Linguistics and the 7th International Joint Conference on Natural Language Processing (Volume 2: Short Papers)}, pages 863--869, Beijing, China. Association for Computational Linguistics.

\bibitem[{Charniak et~al.(2000)Charniak, Blaheta, Ge, Hall, Hale, and Johnson}]{BLLIP-2000}
Eugene Charniak, Don Blaheta, Niyu Ge, Keith Hall, John Hale, and Mark Johnson. 2000.
\newblock Bllip 1987-89 wsj corpus release 1.
\newblock \emph{Linguistic Data Consortium}, 36.

\bibitem[{Chelba(1997)}]{chelba-1997-structured}
Ciprian Chelba. 1997.
\newblock \href {https://doi.org/10.3115/976909.979681} {A structured language model}.
\newblock In \emph{35th Annual Meeting of the Association for Computational Linguistics and 8th Conference of the {E}uropean Chapter of the Association for Computational Linguistics}, pages 498--500, Madrid, Spain. Association for Computational Linguistics.

\bibitem[{Choe and Charniak(2016)}]{choe-charniak-2016-parsing}
Do~Kook Choe and Eugene Charniak. 2016.
\newblock \href {https://doi.org/10.18653/v1/D16-1257} {Parsing as language modeling}.
\newblock In \emph{Proceedings of the 2016 Conference on Empirical Methods in Natural Language Processing}, pages 2331--2336, Austin, Texas. Association for Computational Linguistics.

\bibitem[{Cohen et~al.(2011)Cohen, G{\'o}mez-Rodr{\'\i}guez, and Satta}]{cohen-etal-2011-exact}
Shay~B. Cohen, Carlos G{\'o}mez-Rodr{\'\i}guez, and Giorgio Satta. 2011.
\newblock \href {https://aclanthology.org/D11-1114} {Exact inference for generative probabilistic non-projective dependency parsing}.
\newblock In \emph{Proceedings of the 2011 Conference on Empirical Methods in Natural Language Processing}, pages 1234--1245, Edinburgh, Scotland, UK. Association for Computational Linguistics.

\bibitem[{Dai et~al.(2019)Dai, Yang, Yang, Carbonell, Le, and Salakhutdinov}]{dai-etal-2019-transformer}
Zihang Dai, Zhilin Yang, Yiming Yang, Jaime Carbonell, Quoc Le, and Ruslan Salakhutdinov. 2019.
\newblock \href {https://doi.org/10.18653/v1/P19-1285} {Transformer-{XL}: Attentive language models beyond a fixed-length context}.
\newblock In \emph{Proceedings of the 57th Annual Meeting of the Association for Computational Linguistics}, pages 2978--2988, Florence, Italy. Association for Computational Linguistics.

\bibitem[{Dao et~al.(2022)Dao, Fu, Ermon, Rudra, and R'e}]{Dao2022FlashAttentionFA}
Tri Dao, Daniel~Y. Fu, Stefano Ermon, Atri Rudra, and Christopher R'e. 2022.
\newblock \href {https://api.semanticscholar.org/CorpusID:249151871} {Flashattention: Fast and memory-efficient exact attention with io-awareness}.
\newblock \emph{ArXiv}, abs/2205.14135.

\bibitem[{de~Marneffe and Manning(2008)}]{de-marneffe-manning-2008-stanford}
Marie-Catherine de~Marneffe and Christopher~D. Manning. 2008.
\newblock \href {https://aclanthology.org/W08-1301} {The {S}tanford typed dependencies representation}.
\newblock In \emph{Coling 2008: Proceedings of the workshop on Cross-Framework and Cross-Domain Parser Evaluation}, pages 1--8, Manchester, UK. Coling 2008 Organizing Committee.

\bibitem[{Devlin et~al.(2019)Devlin, Chang, Lee, and Toutanova}]{devlin-etal-2019-bert}
Jacob Devlin, Ming-Wei Chang, Kenton Lee, and Kristina Toutanova. 2019.
\newblock \href {https://doi.org/10.18653/v1/N19-1423} {{BERT}: Pre-training of deep bidirectional transformers for language understanding}.
\newblock In \emph{Proceedings of the 2019 Conference of the North {A}merican Chapter of the Association for Computational Linguistics: Human Language Technologies, Volume 1 (Long and Short Papers)}, pages 4171--4186, Minneapolis, Minnesota. Association for Computational Linguistics.

\bibitem[{Dozat and Manning(2017)}]{dozat2017deep}
Timothy Dozat and Christopher~D. Manning. 2017.
\newblock \href {https://openreview.net/forum?id=Hk95PK9le} {Deep biaffine attention for neural dependency parsing}.
\newblock In \emph{International Conference on Learning Representations}.

\bibitem[{DuSell and Chiang(2021)}]{dusell2021learning}
Brian DuSell and David Chiang. 2021.
\newblock Learning hierarchical structures with differentiable nondeterministic stacks.
\newblock \emph{arXiv preprint arXiv:2109.01982}.

\bibitem[{DuSell and Chiang(2023)}]{dusell2023stack}
Brian DuSell and David Chiang. 2023.
\newblock Stack attention: Improving the ability of transformers to model hierarchical patterns.
\newblock \emph{arXiv preprint arXiv:2310.01749}.

\bibitem[{Dyer et~al.(2016)Dyer, Kuncoro, Ballesteros, and Smith}]{dyer-etal-2016-recurrent}
Chris Dyer, Adhiguna Kuncoro, Miguel Ballesteros, and Noah~A. Smith. 2016.
\newblock \href {https://doi.org/10.18653/v1/N16-1024} {Recurrent neural network grammars}.
\newblock In \emph{Proceedings of the 2016 Conference of the North {A}merican Chapter of the Association for Computational Linguistics: Human Language Technologies}, pages 199--209, San Diego, California. Association for Computational Linguistics.

\bibitem[{Everaert et~al.(2015)Everaert, Huybregts, Chomsky, Berwick, and Bolhuis}]{everaert2015structures}
Martin~BH Everaert, Marinus~AC Huybregts, Noam Chomsky, Robert~C Berwick, and Johan~J Bolhuis. 2015.
\newblock Structures, not strings: Linguistics as part of the cognitive sciences.
\newblock \emph{Trends in cognitive sciences}, 19(12):729--743.

\bibitem[{Fernandez~Astudillo et~al.(2020)Fernandez~Astudillo, Ballesteros, Naseem, Blodgett, and Florian}]{fernandez-astudillo-etal-2020-transition}
Ram{\'o}n Fernandez~Astudillo, Miguel Ballesteros, Tahira Naseem, Austin Blodgett, and Radu Florian. 2020.
\newblock \href {https://doi.org/10.18653/v1/2020.findings-emnlp.89} {Transition-based parsing with stack-transformers}.
\newblock In \emph{Findings of the Association for Computational Linguistics: EMNLP 2020}, pages 1001--1007, Online. Association for Computational Linguistics.

\bibitem[{Hale et~al.(2018)Hale, Dyer, Kuncoro, and Brennan}]{hale-etal-2018-finding}
John Hale, Chris Dyer, Adhiguna Kuncoro, and Jonathan Brennan. 2018.
\newblock \href {https://doi.org/10.18653/v1/P18-1254} {Finding syntax in human encephalography with beam search}.
\newblock In \emph{Proceedings of the 56th Annual Meeting of the Association for Computational Linguistics (Volume 1: Long Papers)}, pages 2727--2736, Melbourne, Australia. Association for Computational Linguistics.

\bibitem[{Henderson(2004)}]{henderson-2004-discriminative}
James Henderson. 2004.
\newblock \href {https://doi.org/10.3115/1218955.1218968} {Discriminative training of a neural network statistical parser}.
\newblock In \emph{Proceedings of the 42nd Annual Meeting of the Association for Computational Linguistics ({ACL}-04)}, pages 95--102, Barcelona, Spain.

\bibitem[{Htut et~al.(2019)Htut, Phang, Bordia, and Bowman}]{Htut2019DoAH}
Phu~Mon Htut, Jason Phang, Shikha Bordia, and Samuel~R. Bowman. 2019.
\newblock Do attention heads in bert track syntactic dependencies?
\newblock \emph{ArXiv}, abs/1911.12246.

\bibitem[{Hu et~al.(2020)Hu, Gauthier, Qian, Wilcox, and Levy}]{hu-etal-2020-systematic}
Jennifer Hu, Jon Gauthier, Peng Qian, Ethan Wilcox, and Roger Levy. 2020.
\newblock \href {https://doi.org/10.18653/v1/2020.acl-main.158} {A systematic assessment of syntactic generalization in neural language models}.
\newblock In \emph{Proceedings of the 58th Annual Meeting of the Association for Computational Linguistics}, pages 1725--1744, Online. Association for Computational Linguistics.

\bibitem[{Joulin and Mikolov(2015)}]{joulin2015inferring}
Armand Joulin and Tomas Mikolov. 2015.
\newblock Inferring algorithmic patterns with stack-augmented recurrent nets.
\newblock \emph{Advances in neural information processing systems}, 28.

\bibitem[{Kim et~al.(2020)Kim, Choi, Edmiston, and goo Lee}]{Kim2020Are}
Taeuk Kim, Jihun Choi, Daniel Edmiston, and Sang goo Lee. 2020.
\newblock \href {https://openreview.net/forum?id=H1xPR3NtPB} {Are pre-trained language models aware of phrases? simple but strong baselines for grammar induction}.
\newblock In \emph{International Conference on Learning Representations}.

\bibitem[{Kim et~al.(2017)Kim, Denton, Hoang, and Rush}]{kim2017structured}
Yoon Kim, Carl Denton, Luong Hoang, and Alexander~M. Rush. 2017.
\newblock \href {https://openreview.net/forum?id=HkE0Nvqlg} {Structured attention networks}.
\newblock In \emph{International Conference on Learning Representations}.

\bibitem[{Kim et~al.(2019)Kim, Rush, Yu, Kuncoro, Dyer, and Melis}]{kim-etal-2019-unsupervised}
Yoon Kim, Alexander Rush, Lei Yu, Adhiguna Kuncoro, Chris Dyer, and G{\'a}bor Melis. 2019.
\newblock \href {https://doi.org/10.18653/v1/N19-1114} {Unsupervised recurrent neural network grammars}.
\newblock In \emph{Proceedings of the 2019 Conference of the North {A}merican Chapter of the Association for Computational Linguistics: Human Language Technologies, Volume 1 (Long and Short Papers)}, pages 1105--1117, Minneapolis, Minnesota. Association for Computational Linguistics.

\bibitem[{Kovaleva et~al.(2019)Kovaleva, Romanov, Rogers, and Rumshisky}]{kovaleva-etal-2019-revealing}
Olga Kovaleva, Alexey Romanov, Anna Rogers, and Anna Rumshisky. 2019.
\newblock \href {https://doi.org/10.18653/v1/D19-1445} {Revealing the dark secrets of {BERT}}.
\newblock In \emph{Proceedings of the 2019 Conference on Empirical Methods in Natural Language Processing and the 9th International Joint Conference on Natural Language Processing (EMNLP-IJCNLP)}, pages 4365--4374, Hong Kong, China. Association for Computational Linguistics.

\bibitem[{Kudo and Richardson(2018)}]{kudo-richardson-2018-sentencepiece}
Taku Kudo and John Richardson. 2018.
\newblock \href {https://doi.org/10.18653/v1/D18-2012} {{S}entence{P}iece: A simple and language independent subword tokenizer and detokenizer for neural text processing}.
\newblock In \emph{Proceedings of the 2018 Conference on Empirical Methods in Natural Language Processing: System Demonstrations}, pages 66--71, Brussels, Belgium. Association for Computational Linguistics.

\bibitem[{Kuhlmann et~al.(2011)Kuhlmann, G{\'o}mez-Rodr{\'\i}guez, and Satta}]{kuhlmann-etal-2011-dynamic}
Marco Kuhlmann, Carlos G{\'o}mez-Rodr{\'\i}guez, and Giorgio Satta. 2011.
\newblock \href {https://aclanthology.org/P11-1068} {Dynamic programming algorithms for transition-based dependency parsers}.
\newblock In \emph{Proceedings of the 49th Annual Meeting of the Association for Computational Linguistics: Human Language Technologies}, pages 673--682, Portland, Oregon, USA. Association for Computational Linguistics.

\bibitem[{Lafferty et~al.(2001)Lafferty, McCallum, and Pereira}]{10.5555/645530.655813}
John~D. Lafferty, Andrew McCallum, and Fernando C.~N. Pereira. 2001.
\newblock Conditional random fields: Probabilistic models for segmenting and labeling sequence data.
\newblock In \emph{Proceedings of the Eighteenth International Conference on Machine Learning}, ICML '01, page 282–289, San Francisco, CA, USA. Morgan Kaufmann Publishers Inc.

\bibitem[{Lou and Tu(2023)}]{lou-tu-2023-amr}
Chao Lou and Kewei Tu. 2023.
\newblock \href {https://doi.org/10.18653/v1/2023.emnlp-main.553} {{AMR} parsing with causal hierarchical attention and pointers}.
\newblock In \emph{Proceedings of the 2023 Conference on Empirical Methods in Natural Language Processing}, pages 8942--8955, Singapore. Association for Computational Linguistics.

\bibitem[{Manning et~al.(2014)Manning, Surdeanu, Bauer, Finkel, Bethard, and McClosky}]{manning-etal-2014-stanford}
Christopher Manning, Mihai Surdeanu, John Bauer, Jenny Finkel, Steven Bethard, and David McClosky. 2014.
\newblock \href {https://doi.org/10.3115/v1/P14-5010} {The {S}tanford {C}ore{NLP} natural language processing toolkit}.
\newblock In \emph{Proceedings of 52nd Annual Meeting of the Association for Computational Linguistics: System Demonstrations}, pages 55--60, Baltimore, Maryland. Association for Computational Linguistics.

\bibitem[{Marcus et~al.(1993)Marcus, Santorini, and Marcinkiewicz}]{marcus-etal-1993-building}
Mitchell~P. Marcus, Beatrice Santorini, and Mary~Ann Marcinkiewicz. 1993.
\newblock \href {https://aclanthology.org/J93-2004} {Building a large annotated corpus of {E}nglish: The {P}enn {T}reebank}.
\newblock \emph{Computational Linguistics}, 19(2):313--330.

\bibitem[{Mirowski and Vlachos(2015)}]{mirowski-vlachos-2015-dependency}
Piotr Mirowski and Andreas Vlachos. 2015.
\newblock \href {https://doi.org/10.3115/v1/P15-2084} {Dependency recurrent neural language models for sentence completion}.
\newblock In \emph{Proceedings of the 53rd Annual Meeting of the Association for Computational Linguistics and the 7th International Joint Conference on Natural Language Processing (Volume 2: Short Papers)}, pages 511--517, Beijing, China. Association for Computational Linguistics.

\bibitem[{Murty et~al.(2023)Murty, Sharma, Andreas, and Manning}]{murty-etal-2023-pushdown}
Shikhar Murty, Pratyusha Sharma, Jacob Andreas, and Christopher Manning. 2023.
\newblock \href {https://doi.org/10.18653/v1/2023.emnlp-main.195} {Pushdown layers: Encoding recursive structure in transformer language models}.
\newblock In \emph{Proceedings of the 2023 Conference on Empirical Methods in Natural Language Processing}, pages 3233--3247, Singapore. Association for Computational Linguistics.

\bibitem[{Nguyen et~al.(2020)Nguyen, Joty, Hoi, and Socher}]{Nguyen2020Tree-Structured}
Xuan-Phi Nguyen, Shafiq Joty, Steven Hoi, and Richard Socher. 2020.
\newblock \href {https://openreview.net/forum?id=HJxK5pEYvr} {Tree-structured attention with hierarchical accumulation}.
\newblock In \emph{International Conference on Learning Representations}.

\bibitem[{Nivre(2003)}]{nivre-2003-efficient}
Joakim Nivre. 2003.
\newblock \href {https://aclanthology.org/W03-3017} {An efficient algorithm for projective dependency parsing}.
\newblock In \emph{Proceedings of the Eighth International Conference on Parsing Technologies}, pages 149--160, Nancy, France.

\bibitem[{Nivre(2004)}]{nivre-2004-incrementality}
Joakim Nivre. 2004.
\newblock \href {https://aclanthology.org/W04-0308} {Incrementality in deterministic dependency parsing}.
\newblock In \emph{Proceedings of the Workshop on Incremental Parsing: Bringing Engineering and Cognition Together}, pages 50--57, Barcelona, Spain. Association for Computational Linguistics.

\bibitem[{Nivre et~al.(2020)Nivre, de~Marneffe, Ginter, Hajivc, Manning, Pyysalo, Schuster, Tyers, and Zeman}]{Nivre2020UniversalDV}
Joakim Nivre, Marie-Catherine de~Marneffe, Filip Ginter, Jan Hajivc, Christopher~D. Manning, Sampo Pyysalo, Sebastian Schuster, Francis~M. Tyers, and Daniel Zeman. 2020.
\newblock Universal dependencies v2: An evergrowing multilingual treebank collection.
\newblock In \emph{International Conference on Language Resources and Evaluation}.

\bibitem[{Peng et~al.(2019)Peng, Schwartz, and Smith}]{peng-etal-2019-palm}
Hao Peng, Roy Schwartz, and Noah~A. Smith. 2019.
\newblock \href {https://doi.org/10.18653/v1/D19-1376} {{P}a{LM}: A hybrid parser and language model}.
\newblock In \emph{Proceedings of the 2019 Conference on Empirical Methods in Natural Language Processing and the 9th International Joint Conference on Natural Language Processing (EMNLP-IJCNLP)}, pages 3644--3651, Hong Kong, China. Association for Computational Linguistics.

\bibitem[{Prange et~al.(2022)Prange, Schneider, and Kong}]{prange-etal-2022-linguistic}
Jakob Prange, Nathan Schneider, and Lingpeng Kong. 2022.
\newblock \href {https://doi.org/10.18653/v1/2022.naacl-main.325} {Linguistic frameworks go toe-to-toe at neuro-symbolic language modeling}.
\newblock In \emph{Proceedings of the 2022 Conference of the North American Chapter of the Association for Computational Linguistics: Human Language Technologies}, pages 4375--4391, Seattle, United States. Association for Computational Linguistics.

\bibitem[{Qi and Manning(2017)}]{qi-manning-2017-arc}
Peng Qi and Christopher~D. Manning. 2017.
\newblock \href {https://doi.org/10.18653/v1/P17-2018} {Arc-swift: A novel transition system for dependency parsing}.
\newblock In \emph{Proceedings of the 55th Annual Meeting of the Association for Computational Linguistics (Volume 2: Short Papers)}, pages 110--117, Vancouver, Canada. Association for Computational Linguistics.

\bibitem[{Qian et~al.(2021)Qian, Naseem, Levy, and Fernandez~Astudillo}]{qian-etal-2021-structural}
Peng Qian, Tahira Naseem, Roger Levy, and Ram{\'o}n Fernandez~Astudillo. 2021.
\newblock \href {https://doi.org/10.18653/v1/2021.acl-long.289} {Structural guidance for transformer language models}.
\newblock In \emph{Proceedings of the 59th Annual Meeting of the Association for Computational Linguistics and the 11th International Joint Conference on Natural Language Processing (Volume 1: Long Papers)}, pages 3735--3745, Online. Association for Computational Linguistics.

\bibitem[{Radford et~al.(2019)Radford, Wu, Child, Luan, Amodei, Sutskever et~al.}]{radford2019language}
Alec Radford, Jeffrey Wu, Rewon Child, David Luan, Dario Amodei, Ilya Sutskever, et~al. 2019.
\newblock Language models are unsupervised multitask learners.
\newblock \emph{OpenAI blog}, 1(8):9.

\bibitem[{Ravishankar et~al.(2021)Ravishankar, Kulmizev, Abdou, S{\o}gaard, and Nivre}]{ravishankar-etal-2021-attention}
Vinit Ravishankar, Artur Kulmizev, Mostafa Abdou, Anders S{\o}gaard, and Joakim Nivre. 2021.
\newblock \href {https://doi.org/10.18653/v1/2021.eacl-main.264} {Attention can reflect syntactic structure (if you let it)}.
\newblock In \emph{Proceedings of the 16th Conference of the European Chapter of the Association for Computational Linguistics: Main Volume}, pages 3031--3045, Online. Association for Computational Linguistics.

\bibitem[{Roark(2001)}]{roark-2001-probabilistic}
Brian Roark. 2001.
\newblock \href {https://doi.org/10.1162/089120101750300526} {Probabilistic top-down parsing and language modeling}.
\newblock \emph{Computational Linguistics}, 27(2):249--276.

\bibitem[{Sartran et~al.(2022)Sartran, Barrett, Kuncoro, Stanojevi{\'c}, Blunsom, and Dyer}]{sartran-etal-2022-transformer}
Laurent Sartran, Samuel Barrett, Adhiguna Kuncoro, Milo{\v{s}} Stanojevi{\'c}, Phil Blunsom, and Chris Dyer. 2022.
\newblock \href {https://doi.org/10.1162/tacl_a_00526} {Transformer grammars: Augmenting transformer language models with syntactic inductive biases at scale}.
\newblock \emph{Transactions of the Association for Computational Linguistics}, 10:1423--1439.

\bibitem[{Shen et~al.(2022)Shen, Tan, Sordoni, Li, Zhou, and Courville}]{shen-etal-2022-unsupervised}
Yikang Shen, Shawn Tan, Alessandro Sordoni, Peng Li, Jie Zhou, and Aaron Courville. 2022.
\newblock \href {https://doi.org/10.18653/v1/2022.acl-long.327} {Unsupervised dependency graph network}.
\newblock In \emph{Proceedings of the 60th Annual Meeting of the Association for Computational Linguistics (Volume 1: Long Papers)}, pages 4767--4784, Dublin, Ireland. Association for Computational Linguistics.

\bibitem[{Shen et~al.(2021)Shen, Tay, Zheng, Bahri, Metzler, and Courville}]{shen-etal-2021-structformer}
Yikang Shen, Yi~Tay, Che Zheng, Dara Bahri, Donald Metzler, and Aaron Courville. 2021.
\newblock \href {https://doi.org/10.18653/v1/2021.acl-long.559} {{S}truct{F}ormer: Joint unsupervised induction of dependency and constituency structure from masked language modeling}.
\newblock In \emph{Proceedings of the 59th Annual Meeting of the Association for Computational Linguistics and the 11th International Joint Conference on Natural Language Processing (Volume 1: Long Papers)}, pages 7196--7209, Online. Association for Computational Linguistics.

\bibitem[{Stern et~al.(2017)Stern, Fried, and Klein}]{stern-etal-2017-effective}
Mitchell Stern, Daniel Fried, and Dan Klein. 2017.
\newblock \href {https://doi.org/10.18653/v1/D17-1178} {Effective inference for generative neural parsing}.
\newblock In \emph{Proceedings of the 2017 Conference on Empirical Methods in Natural Language Processing}, pages 1695--1700, Copenhagen, Denmark. Association for Computational Linguistics.

\bibitem[{Strubell et~al.(2018)Strubell, Verga, Andor, Weiss, and McCallum}]{strubell-etal-2018-linguistically}
Emma Strubell, Patrick Verga, Daniel Andor, David Weiss, and Andrew McCallum. 2018.
\newblock \href {https://doi.org/10.18653/v1/D18-1548} {Linguistically-informed self-attention for semantic role labeling}.
\newblock In \emph{Proceedings of the 2018 Conference on Empirical Methods in Natural Language Processing}, pages 5027--5038, Brussels, Belgium. Association for Computational Linguistics.

\bibitem[{Su et~al.(2021)Su, Lu, Pan, Wen, and Liu}]{Su2021RoFormerET}
Jianlin Su, Yu~Lu, Shengfeng Pan, Bo~Wen, and Yunfeng Liu. 2021.
\newblock \href {https://api.semanticscholar.org/CorpusID:233307138} {Roformer: Enhanced transformer with rotary position embedding}.
\newblock \emph{ArXiv}, abs/2104.09864.

\bibitem[{Titov and Henderson(2007)}]{titov-henderson-2007-latent}
Ivan Titov and James Henderson. 2007.
\newblock \href {https://aclanthology.org/W07-2218} {A latent variable model for generative dependency parsing}.
\newblock In \emph{Proceedings of the Tenth International Conference on Parsing Technologies}, pages 144--155, Prague, Czech Republic. Association for Computational Linguistics.

\bibitem[{Vaswani et~al.(2017)Vaswani, Shazeer, Parmar, Uszkoreit, Jones, Gomez, Kaiser, and Polosukhin}]{vaswani2017attention}
Ashish Vaswani, Noam Shazeer, Niki Parmar, Jakob Uszkoreit, Llion Jones, Aidan~N Gomez, {\L}ukasz Kaiser, and Illia Polosukhin. 2017.
\newblock Attention is all you need.
\newblock \emph{Advances in neural information processing systems}, 30.

\bibitem[{Wang et~al.(2019)Wang, Lee, and Chen}]{wang-etal-2019-tree}
Yaushian Wang, Hung-Yi Lee, and Yun-Nung Chen. 2019.
\newblock \href {https://doi.org/10.18653/v1/D19-1098} {Tree transformer: Integrating tree structures into self-attention}.
\newblock In \emph{Proceedings of the 2019 Conference on Empirical Methods in Natural Language Processing and the 9th International Joint Conference on Natural Language Processing (EMNLP-IJCNLP)}, pages 1061--1070, Hong Kong, China. Association for Computational Linguistics.

\bibitem[{Warstadt et~al.(2020)Warstadt, Parrish, Liu, Mohananey, Peng, Wang, and Bowman}]{warstadt-etal-2020-blimp-benchmark}
Alex Warstadt, Alicia Parrish, Haokun Liu, Anhad Mohananey, Wei Peng, Sheng-Fu Wang, and Samuel~R. Bowman. 2020.
\newblock \href {https://doi.org/10.1162/tacl_a_00321} {{BL}i{MP}: The benchmark of linguistic minimal pairs for {E}nglish}.
\newblock \emph{Transactions of the Association for Computational Linguistics}, 8:377--392.

\bibitem[{Wu and Tu(2023)}]{wu-tu-2023-probabilistic}
Haoyi Wu and Kewei Tu. 2023.
\newblock \href {https://doi.org/10.18653/v1/2023.findings-acl.482} {Probabilistic transformer: A probabilistic dependency model for contextual word representation}.
\newblock In \emph{Findings of the Association for Computational Linguistics: ACL 2023}, pages 7613--7636, Toronto, Canada. Association for Computational Linguistics.

\bibitem[{Yogatama et~al.(2018)Yogatama, Miao, Melis, Ling, Kuncoro, Dyer, and Blunsom}]{yogatama2018memory}
Dani Yogatama, Yishu Miao, Gabor Melis, Wang Ling, Adhiguna Kuncoro, Chris Dyer, and Phil Blunsom. 2018.
\newblock \href {https://openreview.net/forum?id=SkFqf0lAZ} {Memory architectures in recurrent neural network language models}.
\newblock In \emph{International Conference on Learning Representations}.

\bibitem[{Yoshida and Oseki(2022)}]{yoshida-oseki-2022-composition}
Ryo Yoshida and Yohei Oseki. 2022.
\newblock \href {https://doi.org/10.18653/v1/2022.findings-emnlp.428} {Composition, attention, or both?}
\newblock In \emph{Findings of the Association for Computational Linguistics: EMNLP 2022}, pages 5822--5834, Abu Dhabi, United Arab Emirates. Association for Computational Linguistics.

\bibitem[{Zhang et~al.(2020)Zhang, Wu, Zhou, Duan, Zhao, and Wang}]{zhang2020sg}
Zhuosheng Zhang, Yuwei Wu, Junru Zhou, Sufeng Duan, Hai Zhao, and Rui Wang. 2020.
\newblock Sg-net: Syntax-guided machine reading comprehension.
\newblock In \emph{Proceedings of the AAAI Conference on Artificial Intelligence}, volume~34, pages 9636--9643.

\end{thebibliography}

\newpage

\appendix

\section{Examples of \myalgnamesuffix-eager and \myalgnamesuffix-swift}
\label{sec:appendix}

\begin{figure*}[tb]
\centering
\begin{subfigure}[b]{0.47\textwidth}
\centering
\resizebox{\textwidth}{!}{%
\begin{tabular}{c|l|l|l}\toprule
    $i$ & \thead{Input} & \thead{Attn. Mask} & \thead{Label} \\\midrule
    0 & <ROOT> & STACK & \texttt{GEN}(There) \\
    1 & There & STACK & \texttt{GEN}(is) \\
    2 & is & STACK & \texttt{LEFTARC} \\
    3 & \texttt{LEFTARC} + is & COMPOSE & - \\
    4 & \texttt{LEFTARC2} + is & STACK & \texttt{RIGHTARC} \\
    5 & \texttt{RIGHTARC} + <ROOT> & COMPOSE & - \\
    6 & \texttt{RIGHTARC2} + <ROOT> & STACK & \texttt{GEN}(a) \\
    7 & a & STACK & \texttt{GEN}(difference) \\
    8 & difference  & STACK & \texttt{LEFTARC}\\
    9 & \texttt{LEFTARC} + difference & COMPOSE & - \\
    10 & \texttt{LEFTARC2} + difference & STACK & \texttt{RIGHTARC} \\
    11 & \texttt{RIGHTARC} + is &COMPOSE & - \\
    12 & \texttt{RIGHTARC2} + is & STACK & \texttt{POP} \\
    13 & \texttt{POP} & STACK & \texttt{POP} \\
    14 & \texttt{POP} & STACK & \texttt{POP} \\
    15 & \texttt{POP} & STACK & <END> \\\bottomrule
\end{tabular}%
}
\caption{Transition sequence after duplicating \texttt{LEFTARC/RIGHTARC}s. We do not have to make predictions for positions 3, 5, 9, 11.}
\end{subfigure}
\hfill
\begin{subfigure}[b]{0.5\textwidth}
    \centering
    \includegraphics[width=\textwidth,clip, trim=0cm 0cm 0cm 0cm]{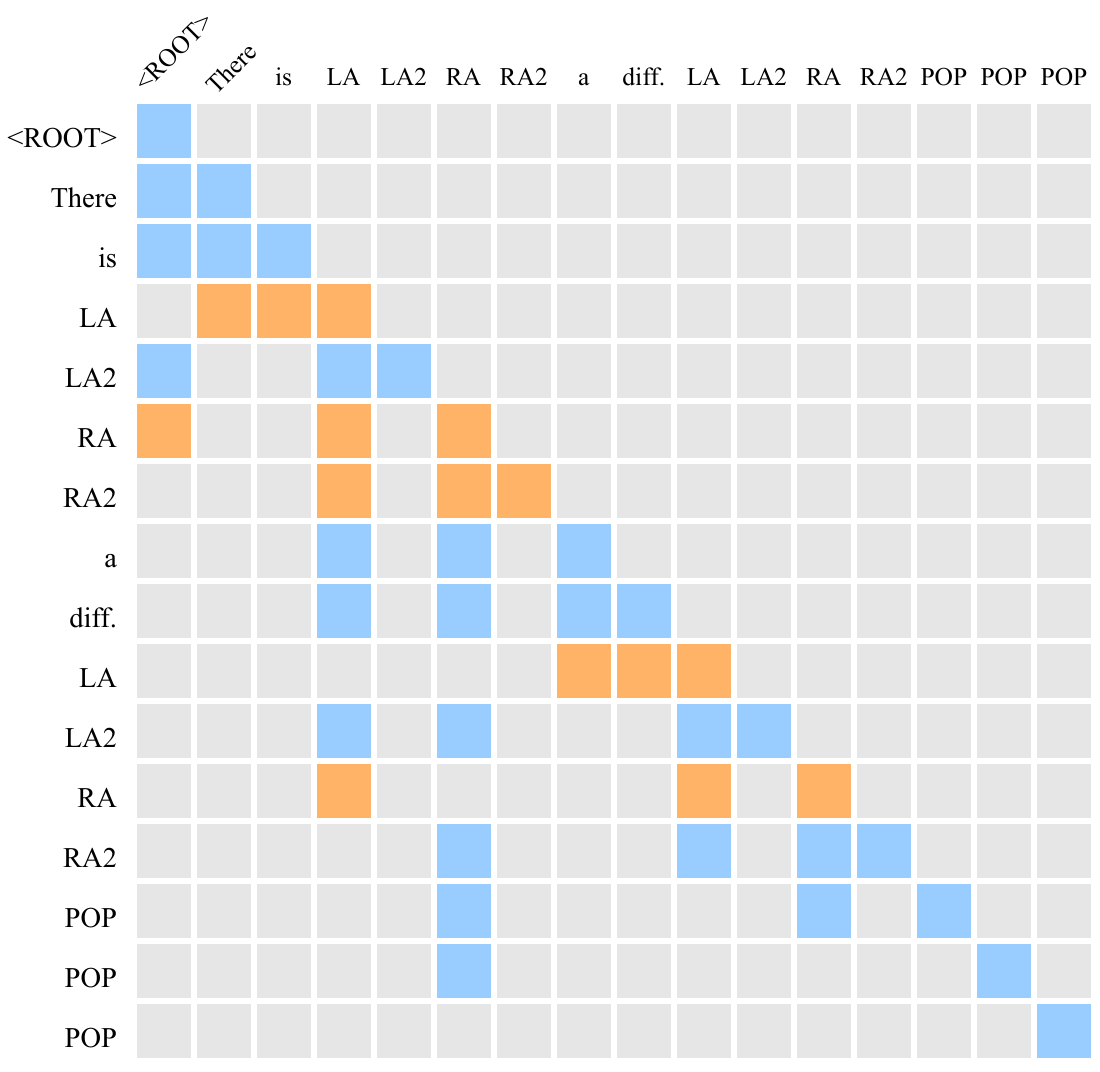}
    \caption{Attention mask. Tokens are simplified for a tight view. We use orange to represent \textbf{COMPOSE} and blue to represent \textbf{STACK} and \textbf{POPSTACK}.}
\end{subfigure}
\caption{Arc-eager processing of an example sentence}
\label{fig:arc_eager_seq_and_mask}
\end{figure*}

The example of \myalgnamesuffix-eager is shown in Figure \ref{fig:arc_eager_seq_and_mask}. The main difference with \myalgname is the new transition \texttt{POP}. It directly masks the top of the stack and attends to other positions, denoted as \textbf{POPSTACK} attention. 

\begin{figure*}[tb]
\centering
\begin{subfigure}[b]{0.47\textwidth}
\centering
\resizebox{\textwidth}{!}{%
\begin{tabular}{c|l|l|l}\toprule
    $i$ & \thead{Input} & \thead{Attn. Mask} & \thead{Label} \\\midrule
    0 & <ROOT> & STACK & \texttt{GEN}(There) \\
    1 & There & STACK & \texttt{GEN}(is) \\
    2 & is & STACK & \texttt{LEFTARC} \\
    3 & \texttt{LEFTARC[1]} + is & COMPOSE & - \\
    4 & \texttt{LEFTARC2[1]} + is & STACK & \texttt{RIGHTARC} \\
    5 & \texttt{RIGHTARC[1]} + <ROOT> & COMPOSE & - \\
    6 & \texttt{RIGHTARC2[1]} + <ROOT> & STACK & \texttt{GEN}(a) \\
    7 & a & STACK & \texttt{GEN}(difference) \\
    8 & difference  & STACK & \texttt{LEFTARC}\\
    9 & \texttt{LEFTARC[1]} + difference & COMPOSE & - \\
    10 & \texttt{LEFTARC2[1]} + difference & STACK & \texttt{RIGHTARC} \\
    11 & \texttt{RIGHTARC[1]} + is &COMPOSE & - \\
    12 & \texttt{RIGHTARC2[1]} + is & STACK & . \\
    13 & . & STACK & \texttt{RIGHTARC} \\
    14 & \texttt{RIGHTARC[2]} + is & COMPOSE & - \\
    15 & \texttt{RIGHTARC2[2]} + is & STACK & <END> \\\bottomrule
\end{tabular}%
}
\caption{Transition sequence after duplicating \texttt{LEFTARC/RIGHTARC}s. We do not have to make predictions for positions 3, 5, 9, 11, 14.}
\end{subfigure}
\hfill
\begin{subfigure}[b]{0.5\textwidth}
    \centering
    \includegraphics[width=\textwidth,clip, trim=0cm 0cm 0cm 0cm]{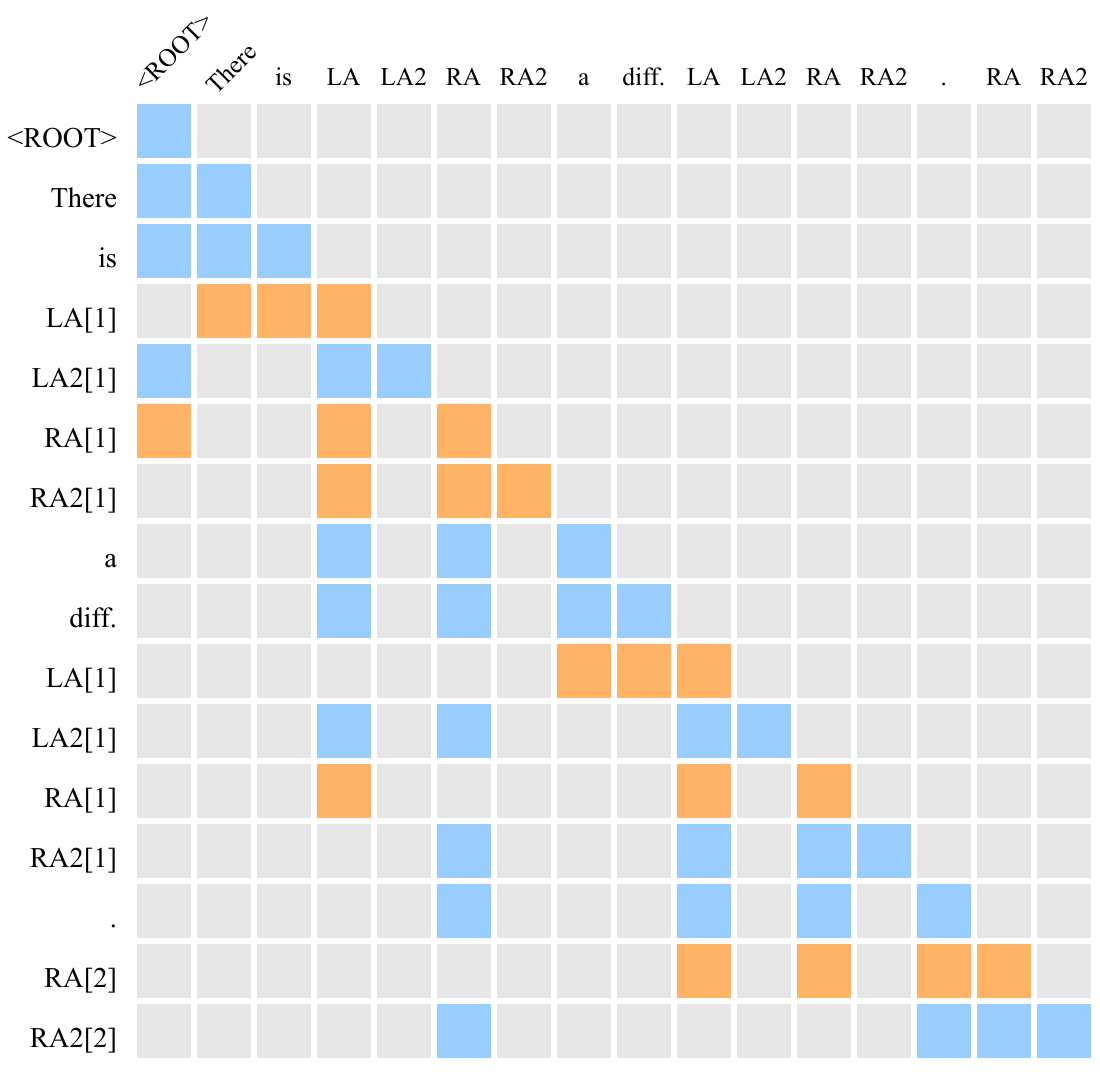}
    \caption{Attention mask. Tokens are simplified for a tight view. We use orange to represent \textbf{COMPOSE} and blue to represent \textbf{STACK}. The number in [] is the \texttt{ARC} number of \texttt{LEFTARC/RIGHTARC}.}
\end{subfigure}
\caption{Arc-swift processing of an example sentence}
\label{fig:arc_swift_seq_and_mask}
\end{figure*}

The example of \myalgnamesuffix-swift is shown in Figure \ref{fig:arc_swift_seq_and_mask}. The newly introduced \texttt{ARC} number is represented within [].

\section{Other Experimental Details}
\label{sec:other_exp_details}
\paragraph{Using subword tokenizers} Previously, we always assume that each word corresponds to a single token. However, subword tokenizers~(e.g., SentencePiece) may divide a word into several subtokens. In our work, we do not consider dependencies among subtokens within a word. All dependencies between words are converted to arcs between the last subtokens of these words. For masking, once a word should be masked, all of its subtokens are masked. For arc representation, we use the embedding of the last subtoken of the head word.

\paragraph{Computational costs} We spent one NVIDIA A6000 GPU for each training, which lasted approximately 35 hours.

\section{Discussion on the Results of BLiMP}
\label{sec:discussion_blimp}
An example testcase in the QUANTIFIERS category of BLiMP is to judge whether ``An actor arrived at at most six lakes'' or ``No actor arrived at at most six lakes'' is acceptable. The correct answer is that the former is acceptable while the latter is not, because superlative quantifiers cannot embed under negation. A stardard Transformer language model could assign a lower probability to the second sentence because it could lower the probability of generating ``at most'' by attending to ``No''. In \myalgnamesingle, however, ``No'' as a determiner is absorbed into ``actor'' and hence masked from the attention when generating ``at most''. While doing this can be beneficial to syntactic generalization, it hinders semantic judgment in this case.

\section{Discussion on the Results of SG}
\label{sec:discussion_SG}
To measure the overlap that DTG and TXL (trans) get right on SG, we count the numbers of correct examples and wrong examples of both models and the results are shown in Table~\ref{tab:Overlap_SG}.

\begin{table}[tb]
    \centering
    \resizebox{\columnwidth}{!}{%
    \begin{tabular}{l|cc} \toprule
        & \thead{DTG correct} & \thead{DTG wrong}  \\\midrule
         \thead{TXL (trans) correct} & $616$ & $57$\\
         \thead{TXL (trans) wrong} & $69$ & $100$\\ \bottomrule    
    \end{tabular}%
    }
    \caption{Numbers of examples that DTG and TXL (trans) get right/wrong on SG.}
    \label{tab:Overlap_SG}
\end{table}

The overlap between the examples that these two models both get right is large. But they still get a few different examples wrong. A representative example is ``The manager to the side of the architects likes to gamble'' and ``The manager to the side of the architects like to gamble''. The former sentence is grammatical while the latter is ungrammatical. TXL (trans) fails to assign a higher surprisal to ``like'' than ``likes'' because it is distracted by the plural form ``architects'' right before the verb ``like'', while DTG manages to do so. The reason for DTG's success can be attributed to the composition of the distracting phrase ``to the side of the architects''. When generating ``likes'', as the distracting phrase has been composed, DTG can focus more on ``The manager''.

\end{document}